\documentclass{egpubl}

 \JournalSubmission    
 \electronicVersion 

\ifpdf \usepackage[pdftex]{graphicx} \pdfcompresslevel=9
\else \usepackage[dvips]{graphicx} \fi

\PrintedOrElectronic

\usepackage{t1enc,dfadobe}

\usepackage{egweblnk}
\usepackage{cite}

\usepackage{amscd} 

\usepackage{amsfonts} 
\usepackage{amsmath} 
\usepackage{amssymb}  

\usepackage[utf8x]{inputenc}

\usepackage{overpic}

\usepackage{bm}
\usepackage{wrapfig}

\renewcommand{\S}{\mathcal{S}}
\newcommand{\M}{\mathcal{M}}
\newcommand{\N}{\mathcal{N}}

\newcommand{\blueround}[1]{\textcolor{blue}{#1}}
\newcommand{\redround}[1]{\textcolor{red}{#1}}
\newcommand{\thename}{\textbf{Ours}}

\title[FARM: Functional Automatic Registration Method for 3D Human Bodies]%
      {FARM:\\ Functional Automatic Registration Method for 3D Human Bodies}

\author[R. Marin et al.]
{\parbox{\textwidth}{\centering R. Marin$^{1}$\thanks{Equal Contribution}, S. Melzi$^{1}$\footnotemark[1],  E. Rodol\`{a}$^{2}$ and U. Castellani$^{1}$
        }
        \\
{\parbox{\textwidth}{\centering 
         $^1$University of Verona  ~~~~~
         $^2$Sapienza University of Rome \\
  }
}
}

\begin{document}


\maketitle
\begin{abstract}
   We introduce a new method for non-rigid registration of 3D human shapes. Our proposed pipeline builds upon a given parametric model of the human, and makes use of the functional map representation for encoding and inferring shape maps throughout the registration process. This combination endows our method with robustness to a large variety of nuisances observed in practical settings, including non-isometric transformations, downsampling, topological noise, and occlusions; further, the pipeline can be applied invariably across different shape representations (e.g. meshes and point clouds), and in the presence of (even dramatic) missing parts such as those arising in real-world depth sensing applications. We showcase our method on a selection of challenging tasks, demonstrating results in line with, or even surpassing, state-of-the-art methods in the respective areas.


\end{abstract}  
\section{Introduction}
\label{sec:introduction}

Non-rigid 3D shape registration is a key problem in computer vision and geometry processing, meeting with increasing attention due to the ever growing amounts of 3D data at our disposal. It is often the case that such data derive from a sensing process, therefore requiring an alignment step in order to fully exploit their informativeness.
The main goal of {\em non-rigid registration} is therefore to determine the correct (according to some task-dependent criterion) non-rigid alignment between two or more data observations. Despite a lot of research being devoted to this issue, however, this problem is far from being solved.

Perhaps the most prominent setting in which non-rigid registration plays a key role is 3D reconstruction of deformable objects. In this context, several partial scans must be aligned non-rigidly to obtain a single object in some canonical pose. This apparently simple task is frustratingly complex due to several reasons; first and foremost, the {\em partial overlap} among the scans as well as the wide variety of {\em noise} factors make this problem particularly challenging. Typical applications include semantic segmentation, motion tracking, recognition, and animation among several others \cite{gleicher1998retargetting,varanasi2008temporal,newcombe2015dynamicfusion}.

The main focus of this paper is non-rigid registration of {\em human} shapes. Despite the less generic setting, we are here confronted with several issues: Human bodies can take countless different poses, there exists a large variety of inter-subject variations (different individuals), and humans interact with the environment giving rise to {\em occlusions, missing parts}, and {\em topological artifacts}. 
In order to address these issues, in this work we make use of a parametric model to which we register the observed data. Our registration method is realized as a full pipeline whose individual steps are carefully designed to maximize accuracy, consistency and robustness, and to avoid any user input. A crucial step of our pipeline relies on {\em functional correspondence}, which enables addressing several challenging forms of artifacts in a unified and consistent language. Importantly, our proposed pipeline is completely automatic, and performs reliably well on a range of challenging cases where other state-of-the-art approaches typically fail.
%



We summarize our \textbf{main contributions} as follows:
\begin{itemize}
\item Our key contribution is a novel {\em fully automatic} pipeline for non-rigid registration of human shapes. To our knowledge, previous approaches either require user input, or impose strong assumptions on the data initialization (e.g., prior alignment).
\item We propose for the first time a {\em unified solution} to address missing parts, topology artifacts, different sampling, occlusions, surface noise, non-isometric transformations, which can be applied invariably to different shape representations including meshes and point clouds.
\item We define a way to identify a set of consistently labelled body landmarks, which is demonstrably robust to the aforementioned types of noise. Additionally, the left/right ambiguity typically found in intrinsically symmetric shapes is completely resolved in the process.
%
\end{itemize}

Finally, we showcase our method on a number of emerging applications in computer vision and geometry processing, demonstrating results that outweigh the state of the art in several challenging settings.

\section{Related work}
\label{sec:related}
Non-rigid surface registration has attracted the attention of several researchers in the last few decades. In order to remain within the scope of our work, we provide here an overview of the methods that are more closely related to our approach.

%
%
%
\vspace{1ex}\noindent\textbf{Non-rigid correspondence.}
%
The literature abounds with fully automatic or semi-automatic methods dealing with sparse or dense correspondence estimation. In \cite{hirshberg2011} the authors proposed a method for registering human bodies under the assumption that the given subjects start with a similar pose; the method exploits face and ankle detection to drive the correspondence process.
In \cite{zhang2017,SPuppet}, a registration method is applied that requires a manual alignment of the human torso; similarly, \cite{Koltun} proposed an optimization procedure based on Markov random fields that assumes the given shapes to be pre-aligned. The method demonstrated high accuracy on a correspondence benchmark of real human shapes (comparisons with this method will be be shown in the experimental section).  

A data-driven approach  for anthropometric landmarking was proposed in \cite{Wuhrer2012} by learning over a large dataset of human shapes in the same pose. 
Differently, in our work we extract stable landmarks over human bodies without the need of data collection, training, or human interaction, since we rely exclusively upon geometric properties in the spectral domain. 
Other purely geometric methods \cite{zhang2008} that work well for human shapes assume the complete absence of topological or geometric errors, limiting their applicability to real-world data.
%
Body landmark detection was recently explored in the SHREC'14 challenge \cite{Giachetti2014}, showing unreliable results under strong changes in pose.

\vspace{2ex}\noindent\textbf{Human body registration.} 
Various model-based techniques have been proposed in the literature. Usually high resolution templates \cite{Allen03} or \emph{morphable models} \cite{Black2012,Anguelov05,loper15} are used to register the target shape. These methods usually start by defining a pose prior under some regularization constraint and sparse correspondence; model and template are then aligned, and shape details are estimated by local non-rigid methods \cite{Igarashi05}. 

Such approaches, however, usually employ accurate hand-placed landmarks. 
%
W\"uhrer et al. \cite{Wuhrer2011} do template fitting based on a dataset of similar shapes; Anguelov et al. \cite{Anguelov2005} enforce the preservation of a constraint over geodesic distances that fails in the presence of topological error and strong isometric distortion. A stochastic approach is given in \cite{SPuppet}, which is based on a random particle system over a segmented template. This method represents the state of the art in the FAUST challenge \cite{FAUST}, but it requires an initialization of the torso to fix the correct body orientation. 
%
Finally, automatic rigging methods like \cite{Baran2007,Feng2014,Feng15} are also related to our approach in that they can be seen as an {\em application} of the registration pipeline. As we will show in the experimental evaluation, automatic rigging for animation is but one of the many tasks that one can address with an automatic registration method at hand.


\section{Background}
\label{sec:background}

\subsection{Continuous surfaces}
\label{sub:continuous}
We model human shapes as two-dimensional Riemannian manifolds $\S$ (possibly with a boundary $\partial \S$) embedded into $\mathbb{R}^{3}$, and equipped with the standard metric induced by the volume form.
%
We denote by $L^2(\S)$ the space of square-integrable real functions on $\S$, and use the standard $L^2(\S)$ inner product $\langle f, g \rangle_{\S} = \int_\S f(x)g(x) dx$.
In analogy to the Laplace operator on flat spaces, the positive semi-definite Laplace-Beltrami operator $\Delta_\S : L^2(\S)\to L^2(\S)$ provides us with the necessary tools to extend Fourier analysis to manifolds. In particular, it admits an eigendecomposition
%
%
 %
 %
%
\begin{equation}\label{eq:lbeig}
{\Delta}_\S \phi_k = \lambda_k \phi_k\,,
\end{equation}
where $0=\lambda_1 \leq \lambda_2 \leq \hdots$ are real eigenvalues, and $\{\phi_k\}_{k\ge1}$ are the corresponding eigenfunctions forming an orthonormal basis of $L^2(\S)$. Any function $f \in L^2(\S)$ can thus be represented via the Fourier series expansion
\begin{eqnarray}
f(x) &=& \sum_{k\geq 1} \langle f, \phi_k \rangle_{\S} \phi_k(x)\,.
\label{eq:fourier}
\end{eqnarray}

\subsection{Functional maps}
\label{sub:fmaps}
Throughout these pages we will be making use of the notion of {\em map} to transport information across surfaces. To address the large variability of data found in practical settings, we look at an especially convenient representation that allows us to model maps {\em compactly} and to infer them {\em robustly}.


Consider two shapes $\M$ and $\N$, and let $\pi: \N\to\M$ be a pointwise map between them. While classical shape matching approaches try to identify point-to-point correspondences (i.e., the map $\pi$) directly, the idea of functional maps~\cite{ovsjanikov2012functional,ovsjanikov16} is to consider a linear operator $T: L^2(\M)\to L^2(\N)$ mapping functions on $\M$ to functions on $\N$, defined as the composition $T(f) = f \circ \pi$. In the Laplace-Beltrami eigenbasis, the operator $T$ (henceforth {\em functional map}) admits a matrix representation $\mathbf{C}=(c_{ij})$, with coefficients defined according to:
\begin{equation}\label{eq:c}
T(f) = T \sum_{i} \langle f, \phi_i \rangle_\M \phi_i = \sum_{ij} \langle f, \phi_i \rangle_\M \underbrace{\langle T\phi_i , \psi_j \rangle_\N}_{c_{ji}}\,,
\end{equation}
where $\{\phi_i\}$ and $\{\psi_j\}$ are the Laplacian eigenbases on $\M$ and $\N$ respectively. As suggested in~\cite{ovsjanikov2012functional}, the series \eqref{eq:c} can be truncated after the first $k$ coefficients, yielding a band-limited approximation (in the Fourier sense) of the underlying map $\pi$. Estimating a functional map in the Fourier basis thus boils down to solving for a matrix $\mathbf{C}\in\mathbb{R}^{k \times k}$, as opposed to the classical full (and usually binary) correspondence matrix ${\bm{\Pi}}\in\mathbb{R}^{n\times n}$ (here $n$ is the number of surface points in the discrete setting), where typically $k \ll n$.

\begin{figure*}[t]
\begin{overpic}
  [trim=0cm 0cm 0cm 0cm,clip,width=\linewidth]{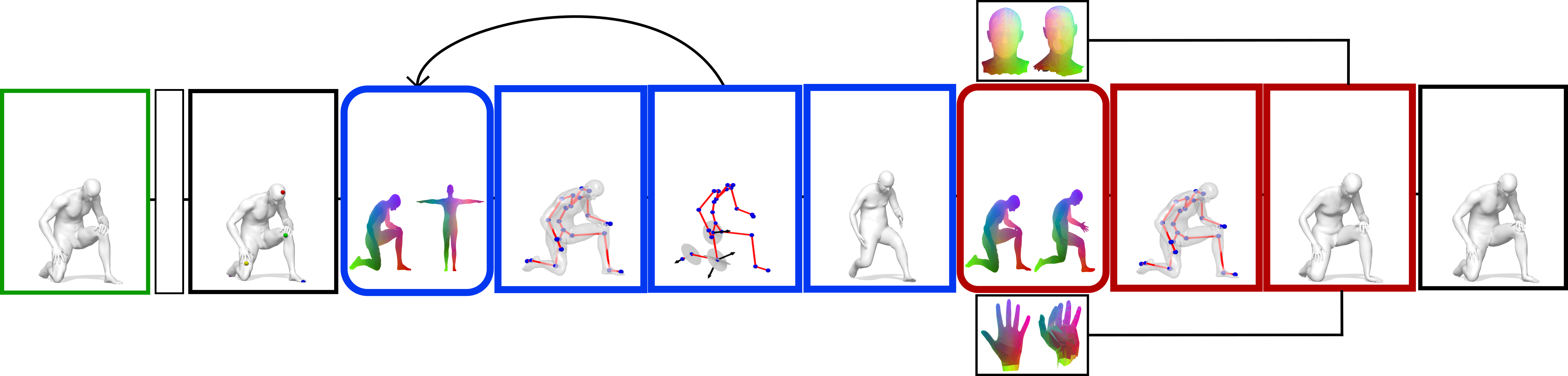}
 \put(2.3, 14.5){\textbf{Target}}
 \put(10.06, 6.7){\rotatebox{90}{preprocessing}}
 
 \put(13, 15.5){Landmarks}
 \put(13.3, 13.5){placement}
 
 \put(25.2, 15.5){{FM}}
 \put(23.9, 13.5){{+ refine}} 
 
 \put(32.3, 15.5){Functional}
 \put(33.5, 13.5){skeleton}
 
 \put(42.2, 15.5){Left/Right}
 \put(43.5, 13.5){labeling}

 \put(52.8, 15.5){{Non-rigid}}
 \put(55.0, 13.5){{ICP}}
 
 \put(64.4, 15.5){{FM}}
 \put(63.3, 13.5){{+ refine}}
 
 \put(71.75, 15.5){{Functional}}
 \put(73.1, 13.5){{skeleton}}
 
 \put(81.9, 15.5){{Non-rigid}}
 \put(83.85, 13.5){{ICP}} 
 
 \put(60.5, 19.2){\rotatebox{90}{{Head}}}
 \put(60.5, 0){\rotatebox{90}{{Hands}}}

 \put(38, 3.5){\textbf{\blueround{Round 1}}}
 \put(72.5, 3.5){\textbf{\redround{Round 2}}}

 \put(93.5, 15.5){{Local}}
 \put(91.5, 13.5){{refinement}} 
 
 \put(55, 3){\textbf{R1}}
 \put(94.5, 3){\textbf{R2}}

\end{overpic}
\vspace{-0.4cm}
\caption{\label{fig:pipeline} Our registration pipeline. We refer to the main text for details on the individual steps. To get a sense of the results, compare the {\em \textbf{Target}} shape with the shapes in boxes {\em \textbf{R1}} and {\em \textbf{R2}}. See also Figure~\ref{fig:rounds}.}
\end{figure*}

\begin{figure}[tb]
\begin{center}
\begin{overpic}
  [trim=0cm 0cm 0cm 0cm,clip,width=\linewidth]{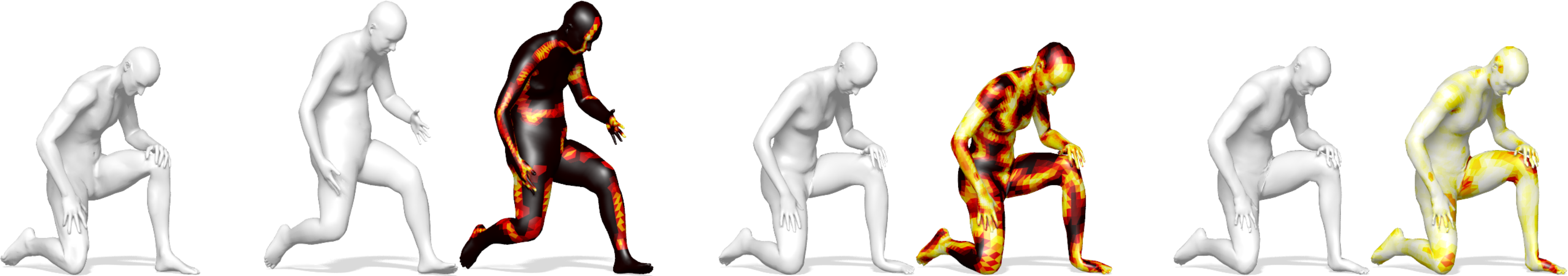}
 \put(3, -3){\footnotesize Target}
 \put(25, -3){\footnotesize Round 1}
 \put(54, -3){\footnotesize Round 2}
 \put(76, -3){\footnotesize Local refinement}
\end{overpic}
\end{center}

\vspace{0.1cm}

\caption{\label{fig:rounds} Registration results after Round 1, Round 2 and local refinement. The heatmap encodes point-to-surface
registration error (expressed in cm, saturated at a maximum value of 1).}
\end{figure}
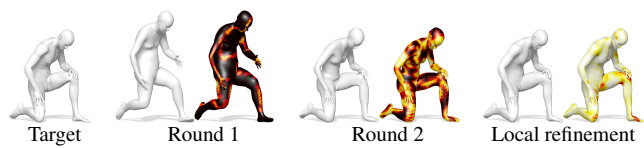

The functional map representation has been successfully used in recent years to estimate dense correspondence between deformable 3D shapes \cite{ovsjanikov2012functional,pokrass13}, in the presence of missing parts or clutter \cite{rodola2016partial,litany16,cosmo2016matching,litany2017fully}, as well as in machine learning pipelines~\cite{corman2014supervised,FMNET}.
In this paper we leverage such flexibility to address several challenging registration scenarios in a unified manner.

\vspace{2ex}\noindent\textbf{Remark. }
By embracing the functional map representation, we shift the difficulty of accounting for geometry and partiality artifacts from the embedding to the functional space, which has a vector space structure, thus allowing us to operate completely within the realm of linear algebra.

\subsection{Parametric model}
\label{sub:SMPL}
Our registration pipeline employs a parametric model for the human body~\cite{Allen03,Anguelov05}, which is to be fitted to a given, possibly very noisy and deformed input observation. In this paper we adopt SMPL~\cite{loper15}, a skinned vertex-based model for human shapes learned over the CAESAR dataset. 
Our choice is mainly motivated by its relatively small number of parameters; together with the functional map representation, this choice endows our approach with desirable efficiency and representation compactness. 
To demonstrate the flexibility of our pipeline, in the experimental section we additionally show results with an alternative parametric model~\cite{pishchulin15arxiv}.

SMPL allows to model both shape (i.e., different subjects) and pose, exposing for this purpose two sets of parameters, namely: (1) shape parameters ${\bm{\beta}}\in\mathbb{R}^{10}$, representing the variation of bodily characteristics in a population of individuals, obtained through learned PCA; and (2) pose parameters $\bm{\theta}\in\mathbb{R}^{72}$, encoding the relative rotation of each of 24 joints with respect to its parent in the kinematic tree, using axis-angle notation.
The SMPL toolset additionally comes with a joint regressor, namely a linear mapper from the 3D vertices of a low-resolution, fixed human template to the 3D coordinates of the 24 skeleton joints. While we make use of this simple regressor in our pipeline, we stress that any other skeleton estimation technique \cite{tagliasacchi2009curve} can be used for this step.

\subsection{Discretization}
\label{sub:discrete}
In the discrete setting, we represent each surface $\S$ as a triangular mesh $(\mathcal{V}, \mathcal{E}, \mathcal{F})$, where $\mathcal{V}$ is a set of vertices sampled over the 3D embedding of the surface, connected by undirected edges $\mathcal{E}$ so as to form a mesh of triangular faces $\mathcal{F}$.
Scalar functions $f:\S\to\mathbb{R}$ are discretized as $|\mathcal{V}|$-dimensional vectors $\mathbf{f}$, where each entry corresponds to the value of the function at the corresponding vertex; following standard practice, we assume functions to behave linearly within each triangle. Bivariate functions $d(x,y):\S\times\S\to\mathbb{R}$ are discretized as $|\mathcal{V}|\times |\mathcal{V}|$ matrices $\mathbf{D}$.

The Laplace-Beltrami operator on $\S$ takes the form of a $|\mathcal{V}|\times |\mathcal{V}|$ matrix $\bm{\Delta}_{\S} = \mathbf{A}^{-1}\mathbf{W}$, where $\mathbf{A} = \mathrm{diag}(a_1, \hdots, a_{|\mathcal{V}|})$ is a diagonal matrix of local area elements, and $\mathbf{W}$ is a matrix of cotangent weights \cite{pinkall93}. On point clouds, $\bm{\Delta}_{\S}$ is discretized via local Delaunay fitting as in \cite{boscaini2016anisotropic}.
Manifold inner products are discretized as $\left\langle \mathbf{f}, \mathbf{g} \right\rangle_\S = \mathbf{f}^\top \mathbf{A} \mathbf{g}$; similarly, application of an integral operator mapping $f(x) \mapsto \int_\S k(x,y) f(y) dy$, defined upon a bivariate kernel $k(\cdot,\cdot)$, is discretized as the product $\mathbf{KAf}$.

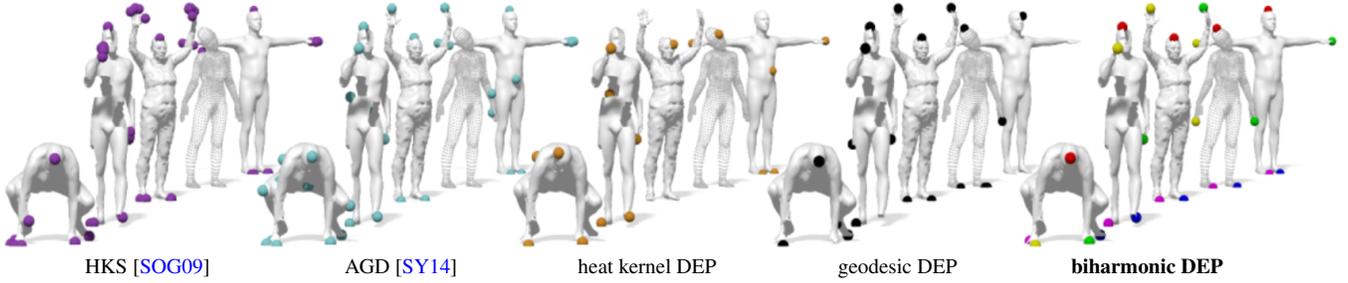
\begin{figure*}[t]
\begin{overpic}[trim=0cm 0cm 0cm 0cm,clip,width=\linewidth]{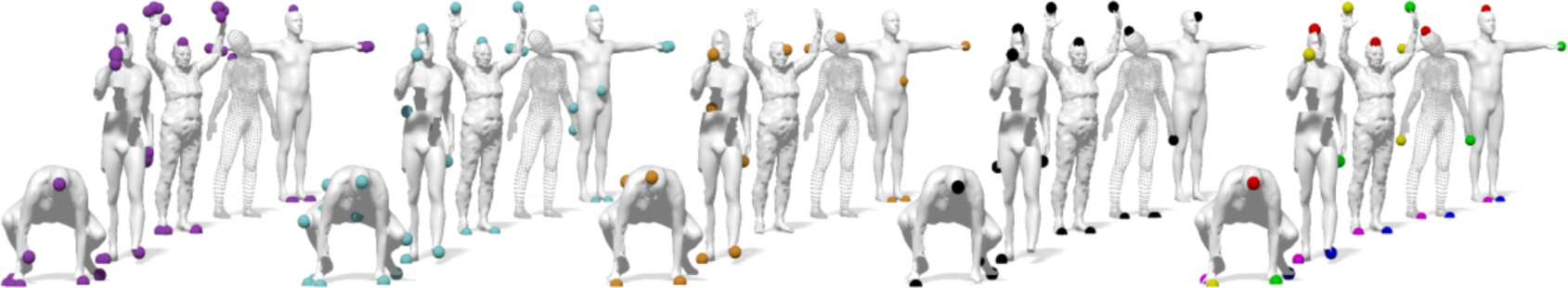}
\put(6,-2){\footnotesize HKS \cite{sun2009concise}}
\put(25.5,-2){\footnotesize AGD \cite{sahilliouglu2014partial}}
\put(43,-2){\footnotesize heat kernel DEP}
\put(62.5,-2){\footnotesize geodesic DEP}
\put(80,-2){\footnotesize \textbf{biharmonic DEP}}
\end{overpic}
\vspace{0.01cm}
\caption{\label{fig:landmarks}Landmarks stability under mesh perturbation. We compare our approach with heat diffusion~\cite{sun2009concise}, a geodesics-based approach~\cite{sahilliouglu2014partial}, and different kernel choices for the score function \eqref{eq:score}. Our proposed solution (biharmonic DEP) returns stable head/hands/feet landmarks under topological gluing, missing parts, surface noise, point cloud representation, and clean meshes (left-to-right within each mesh sequence).}
\end{figure*}

\section{Proposed method}
\label{sec:method}

We present the steps of our method as separate modules, which are then composed in a full registration pipeline.
We emphasize here that our approach is completely automatic as {\em it requires no human supervision}; this is in contrast with a number of existing state-of-the-art approaches, whose initialization either relies on a set of sparse hand-picked matches, or on the assumption that the given human shapes are placed in approximate rigid alignment. A direct comparison with such approaches, with and without human supervision, will be provided in Section~\ref{sec:results}. The overall pipeline is illustrated in Figure~\ref{fig:pipeline}.

The complete code for our method will be made publicly available upon acceptance.


\subsection{Landmarks}
\label{sub:landmarks}
This module consists in identifying and labeling a sparse set of body landmarks for a given input 3D model. These landmarks are used to drive the matching process in the subsequent steps; importantly, since our landmark extraction procedure is resilient to noise, partiality, and topological artifacts, it allows addressing several challenging cases that may arise in a practical setting.

\vspace{1ex}\noindent\textbf{Score function.}
Landmark placement is based upon the construction of a discrete-time evolution process (DEP) \cite{bestpaper} on the mesh surface, realized by defining the recursive relations:
\begin{equation}\label{eq:evo}
f_{(t+1)} = A f_{(t)}
\end{equation}
for scalar functions $f_{(t)}:\S\to\mathbb{R}$ and an integral operator defined by the action
\begin{equation}\label{eq:af}
Af_{(t)} = \int_\S d(\cdot,y) f_{(t)}(y) dy \,,
\end{equation}
where $d:\S\times\S\to\mathbb{R}_+$ is a pairwise potential that depends on the underlying geometry of the surface; if available, one may consider a color-based potential $d:\mathcal{C}(\S)\times \mathcal{C}(\S)\to\mathbb{R}_+$, where $\mathcal{C}(\S)$ is a texture map for surface $\S$. Intuitively, function $d$ encodes the degree of influence that surface points exert on each other, and its selection is crucial for achieving robustness to different types of artifacts.

For a fixed number $T$ of time steps, we consider the score:
\begin{equation}\label{eq:score}
s(x) = f_0(x) + \sum_{t=1}^{T} A^t f_0(x)\,,
\end{equation}
summing up the contributions of the evolution process \eqref{eq:evo} across all discrete times $t=1,\dots,T$. Here $A^t$ denotes repeated application $A(A(\cdots(A)))$ of the operator $t$ times.
A DEP descriptor is obtained by letting $T \to \infty$ and using a multiscale approach on the choice of the pairwise potential, as shown below.

%
%
%
%

\vspace{1ex}\noindent\textbf{Pairwise potential.}
In this paper we advocate the adoption of biharmonic distances \cite{lipman2010biharmonic}, 
%
%
%
%
%
 due to their efficiency and robustness to missing parts and resampling. When used in the definition of the score, they lead to observed resilience to inter- and intra-subject variation, partiality, surface noise and topological gluing.
Our complete pairwise potential is defined as:
\begin{equation}
[0,1] \ni d(x,y) = 1 - \frac{d_B^{\tau}(x,y)}{\mathrm{diam}_B(\S)}\,, 
\end{equation}
\begin{equation}
\mathrm{where} \ \ d_B^\tau (x,y) = 
\begin{cases} 
    d_B(x,y) & d_B(x,y)\le\tau \\
     1 & \mathrm{otherwise} 
   \end{cases} \nonumber
\end{equation}
%
%
and $\mathrm{diam}_B(\S) \equiv \max_{x,y\in\S} d_B(x,y)$ is the biharmonic diameter of surface $\S$. 
The thresholding operation makes $d_B^\tau$ more local, thus bringing increased resilience to partiality and topological noise.

\vspace{1ex}\noindent\textbf{Landmark extraction.}
We use a constant initial state $f_{(0)}(x)=1 ~\forall x\in\S$ and distance thresholds $\tau_1 =0.05, \tau_2=1$, resulting in two score functions $s^{\tau_1}, s^{\tau_2}$ (depicted in the inset).
\setlength{\columnsep}{0pt}
\setlength{\intextsep}{0pt}
\begin{wrapfigure}[13]{r}{0.14\textwidth}
\begin{center}
\begin{overpic}[trim=0cm 0cm 0cm 0cm,clip,width=0.88\linewidth]{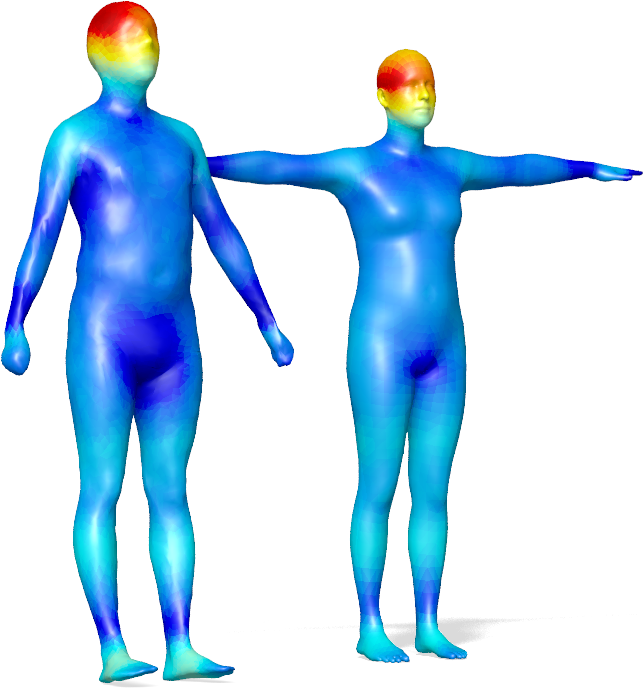}
\put(80,85){$s^{\tau_1}$}
\end{overpic}
\vspace{0.1cm}
\begin{overpic}[trim=0cm 0cm 0cm 0cm,clip,width=0.88\linewidth]{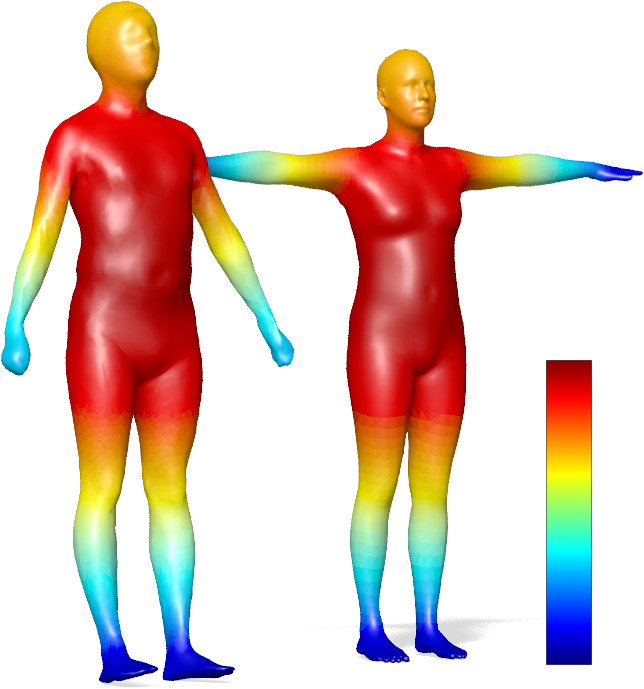}
\put(80,85){$s^{\tau_2}$}
\put(88,2){\tiny 0}
\put(88,43){\tiny 1}
\end{overpic}
\end{center}
\end{wrapfigure}
We mark the tip of the head by seeking for a local (within the region identified by $s^{\tau_1}$) extremum of the first 5 non-constant Laplacian eigenfunctions; $s^{\tau_1}$ is observed to reliably correspond to the head region, while the eigenfunction extrema tend to concentrate around shape protrusions.
The remaining landmarks are identified by considering the 4 clusters of points having a value of $s^{\tau_2}$ below $0.9$. For each cluster, we keep the point that is farther from the head, resulting in 4 unlabeled landmarks. The hand/foot labels are assigned according to the distance to the head landmark. 

We remark that at this point, although we are able to determine the correct hand/foot pairings according to the side of the body they reside in, we are {\em not yet} able to attach a semantic left/right labeling for them; we will do this in the following sections. See Figure~\ref{fig:landmarks} for an evaluation of landmark placement.

%




\subsection{Map inference}
\label{sub:fmaps_norm21}
Registering deformable surfaces entails the computation of dense maps as an intermediate step in the alignment process. We adopt the functional map representation in the Laplacian eigenbasis (Section~\ref{sub:fmaps}), due to the guaranteed invariance to isometric transformations (changes in pose), resilience to mesh downsampling, applicability to different representations (e.g., meshes vs point clouds), surface noise, and compactness of the resulting map representation. Further, functional maps can be robustly estimated in the presence of missing parts, clutter, and alterations of the mesh topology (e.g., ``gluing'' of the discrete surface around areas of self-contact). To our knowledge, there are no other methods allowing to address this variety of issues in a unified language.

\vspace{2ex}\noindent\textbf{Estimating a functional map.}
Let $\M$ be a fixed template (with $n_\M$ vertices) in a canonical pose, and let $\N$ be the observed, possibly noisy and incomplete data (with $n_\N$ vertices). We estimate a functional map $\mathbf{C}$ between $L^2(\M)$ and $L^2(\N)$ as the solution to the following non-convex problem:
\begin{align}\label{eq:comm}
\min_{\mathbf{C}}  \| \mathbf{C} \hat{\mathbf{F}} - \hat{\mathbf{G}}\mathbf{C} \|_F^2 + \lambda_1\| \mathbf{C}\hat{\mathbf{F}} - \hat{\mathbf{G}} \|_F^2 + \lambda_2 \| \mathbf{C}{\bm{\Lambda}}_\M - {\bm{\Lambda}}_\N \mathbf{C} \|_F^2\,
\end{align}
where $\mathbf{C}\in\mathbb{R}^{k_\N\times k_\M}$ is the functional map expressed in the Laplacian eigenbases ${\bm{\Phi}}\in\mathbb{R}^{n_\M\times k_\M},{\bm{\Psi}}\in\mathbb{R}^{n_\N\times k_\N}$, and ${\bm \Lambda}_\M\in\mathbb{R}^{k_\M\times k_\M}, {\bm \Lambda}_\N\in\mathbb{R}^{k_\N\times k_\N}$ are diagonal matrices of the Laplacian eigenvalues. Matrices $\hat{\mathbf{F}}\in\mathbb{R}^{k_\M\times q},\hat{\mathbf{G}}\in\mathbb{R}^{k_\N\times q}$ contain the Fourier expansion coefficients of $q$ probe functions $f_i:\M\to\mathbb{R},g_i:\N\to\mathbb{R},i=1,\dots,q$, i.e., $(a_{ij}) = \langle \phi_i, f_j\rangle_\M, (b_{ij}) = \langle \psi_i, g_j\rangle_\N$.

Problem \eqref{eq:comm} allows to estimate functional maps in a considerably more accurate way than the baseline approach of \cite{ovsjanikov2012functional}. We refer to \cite{Nogneng2017} for details on the motivations behind this energy -- the main rationale being that the commutativity penalty $\| \mathbf{C}\hat{\mathbf{F}} - \hat{\mathbf{G}}\mathbf{C} \|$ promotes solutions that more closely resemble pointwise maps. 

A local optimum to \eqref{eq:comm} is obtained via conjugate gradient, and further refined with the spectral ICP-like method of \cite{ovsjanikov2012functional}. In all our tests we used $k_\M=50, k_\N=30$, and $\lambda_1=0.1, \lambda_2=0.001$ (default values used in \cite{Nogneng2017}). As probe functions $f_i,g_i$, for the first step we use 20-dimensional WKS descriptors~\cite{aubry2011wave} concatenated with 20-dimensional wave kernel maps~\cite{ovsjanikov2012functional} around each body landmark.

\vspace{2ex}\noindent\textbf{Conversion to pointwise map.}
Given a functional map matrix $\mathbf{C}$, the underlying pointwise map ${\bm{\Pi}}\in\{0,1\}^{n_\N\times n_\M}$ is recovered by solving the recovery problem~\cite{ovsjanikov16}
\begin{align}\label{eq:convert}
\min_{\bm{\Pi}} \| \mathbf{C} {\bm{\Phi}}^\top - {\bm{\Psi}}^\top {\bm{\Pi}} \|_F^2 \quad \mathrm{s.t.}~{\bm{\Pi}}^\top\mathbf{1}=\mathbf{1}\,.
\end{align}
If the underlying map is bijective, we would expect the matrix $\bm{\Pi}$ to be a permutation; however, for increased flexibility (e.g., to allow addressing partiality) we relax this constraint to left-stochasticity. We then solve the problem above {\em globally} by a nearest-neighbor approach akin to~\cite{ovsjanikov2012functional}.

\vspace{2ex}\noindent\textbf{Refinement.}
The goal of this module is to improve the quality of an input map by filtering out gross mismatches. We do so by considering a sequence of {\em convex} problems:
\begin{align}\label{eq:refine}
\mathbf{C}^{(t+1)}=\arg\min_{\mathbf{C}} \| \mathbf{C}^{(t)}\hat{\mathbf{F}}^{(t)} - \hat{\mathbf{G}}^{(t)} \|_{2,1} + \mu \| \mathbf{C}^{(t)}\circ \mathbf{W} \|_F^2\,,
\end{align}
with $t=0,\dots,T$ and $\mathbf{C}^{(0)}$ being the input map to refine. 
If the input map has a pointwise representation ${\bm{\Pi}}^{(0)}$, it is first converted to a spectral representation by the change of basis $\mathbf{C}^{(0)} = {\bm{\Psi}}^\top \mathbf{A} {\bm{\Pi}}^{(0)} {\bm{\Phi}}$. 
\setlength{\columnsep}{0pt}
\setlength{\intextsep}{0pt}
\begin{wrapfigure}[8]{l}{0.25\textwidth}
\begin{center}
\begin{overpic}[trim=0cm 0cm 0cm 0cm,clip,width=0.88\linewidth]{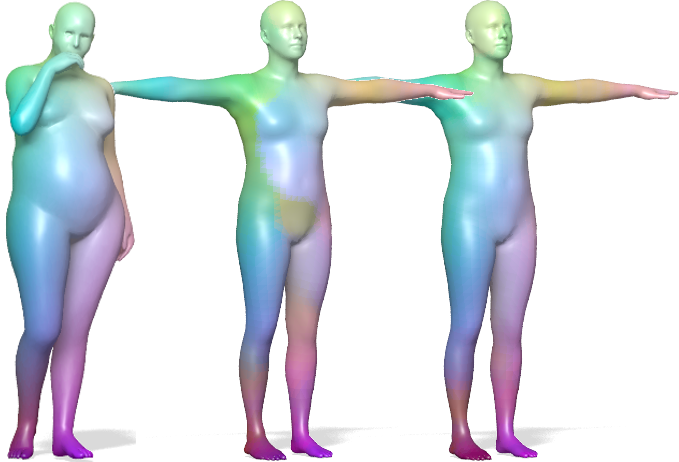}
\put(2,-7){\footnotesize target}
\put(36,-7){\footnotesize noisy}
\put(66,-7){\footnotesize refined}
\end{overpic}
\end{center}
\end{wrapfigure}
The $\mu$-term enforces a diagonal structure on matrix $\mathbf{C}$, where the shape of the diagonal is encoded in the ``mask'' matrix $\mathbf{W}$; this allows to address {\em partiality} by simply setting the diagonal angle of $\mathbf{W}$ according to the area ratio $\frac{\mathrm{area}(\N)}{\mathrm{area}(\M)}$~\cite{rodola2016partial}. 
An example of map refinement is shown in the inset (corresponding points between target and model have same color).

\vspace{1ex}\noindent\textbf{Remark. }Map refinement works {\em as-is} under missing geometry and topological noise, as we will demonstrate in the experiments.

Here, as probe functions $(f_i, g_i)_{i=1}^q$ we use pairs of deltas $(\delta^\M_{x_i}(x), \delta^\N_{\pi^{(0)}(x_i)}(y))_{i=1}^q$ supported at corresponding points $(x_i, \pi^{(0)}(x_i))_{i=1}^q$ where the map $\pi^{(0)}$ is the one given as input. Input functional maps $\mathbf{C}^{(0)}$ are converted to ${\bm{\Pi}}^{(0)}$ by solving \eqref{eq:convert}.

A crucial element of this refinement step is the adoption of the $\ell_{2,1}$ norm in the data term of \eqref{eq:refine}. The norm $\|\mathbf{A}\|_{2,1}$ promotes column-wise sparsity for matrix $\mathbf{A}$; in our setting, it is exactly this type of sparsity that allows to filter out mismatches in the input (recall that our probe functions, which are organized as columns of $\hat{\mathbf{F}},\hat{\mathbf{G}}$, are deltas supported at the input matches).

In all our tests, we used $\mu=0.01$, $T=5$ iterations, and $q=1000$ delta functions supported at uniformly distributed points over $\M$.


\subsection{Left/Right labeling}
\label{sub:simmetry}
\begin{figure}[b]
\begin{center}
\begin{overpic}[trim=0cm 0cm 0cm 0cm,clip,width=\linewidth]{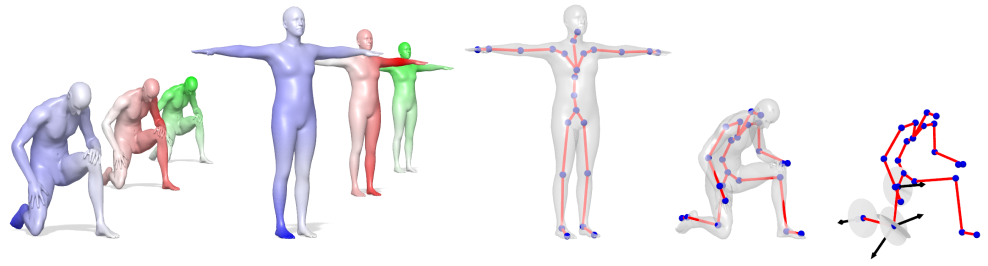}
\put(21,-1){\footnotesize (a)}
\put(20,6.5){\footnotesize $\to$}
\put(4,20){\footnotesize $f_x$}
\put(11,20){\footnotesize $f_y$}
\put(15.5,20){\footnotesize $f_z$}
\put(57,-1){\footnotesize (b)}
\put(73.5,-1){\footnotesize (c)}
\put(92.5,-1){\footnotesize (d)}
\end{overpic}
\vspace{-0.4cm}
\caption{\label{fig:skeleton}(a) The vertex coordinate functions are mapped from shape to template via an estimated functional map; (b) a joint regressor is defined on the template, and (c) it is applied to the mapped coordinates to obtain a skeleton for the shape; (d) the front-facing direction is given by transporting the foot versor up to the rest of the body. This entire sequence is completely automatic.}
\end{center}
\end{figure}
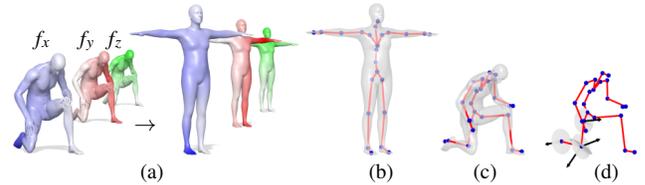

%
%
%
%
Resolving the left/right ambiguity typical of intrinsic methods is crucial for a successful registration pipeline.
%
%
%
%
To this end, the body landmarks are first used to solve for a low-rank functional map $\mathbf{C}$ between the parametric template $\M$ and the input shape $\N$; this is done by solving problem \eqref{eq:comm}. The coordinate functions of $\N$ (i.e., three scalar functions $f_x,f_y,f_z:\N\to\mathbb{R}$ encoding the $x,y,z$ vertex coordinates of $\N$) are then mapped onto $\M$ via $\mathbf{C}$. Note that for the transport of functions a full point-to-point map is not necessary, and indeed a low-rank functional map suffices. A joint regressor is finally used on the mapped coordinates over $\M$, obtaining the skeleton for $\N$ (see Figure~\ref{fig:skeleton}).

Note that, since the body landmarks do {\em not} at this point contain the left/right information, the estimated map might be either the correct one or its symmetrically flipped counterpart. In order to determine which is the case, we detect the front/back symmetry by declaring the tip of the feet (whose landmarks are at our disposal) to be front-facing, and propagate the associated versor up to the rest of the body under torque-penalizing constraints (Figure~\ref{fig:skeleton} rightmost column; for a detailed algorithm we refer to the supplementary material). The front-facing direction, together with the semantic information attached to the parametric skeleton, can then be used to attribute the correct left/right labels to the landmarks.



\begin{figure*}[t]
\begin{center}
\begin{overpic}
  [trim=0cm 0cm 0cm 0cm,clip,width=\linewidth]{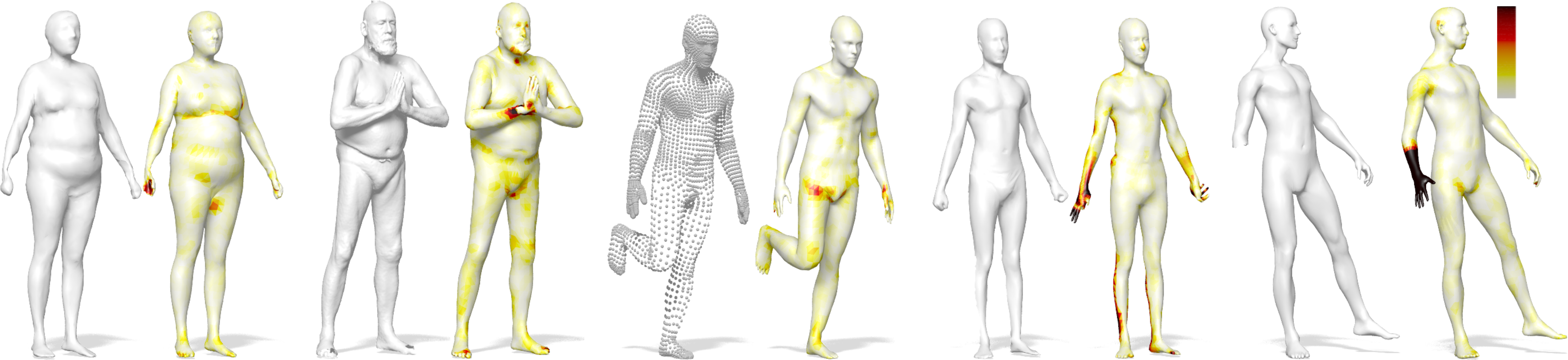}
	\put(4.5, -3){sub-sampled}
	\put(21.5, -3){topological error}
	\put(43.5, -3){point cloud}
	\put(62, -3){partial frontal view}
	\put(85, -3){missing parts}
	\put(97, 21.85){\tiny 1 (cm)}
	\put(97, 16.2){\tiny 0}
	
\end{overpic}
\vspace{0.2cm}
\caption{\label{fig:challenging}Registration results in different settings. We plot the target surface on the left and the registered parametric model on the right. Here and for the rest of the paper, the heatmap encodes point-to-surface registration error (expressed in cm, saturated at 1).
}
\end{center}
\end{figure*}
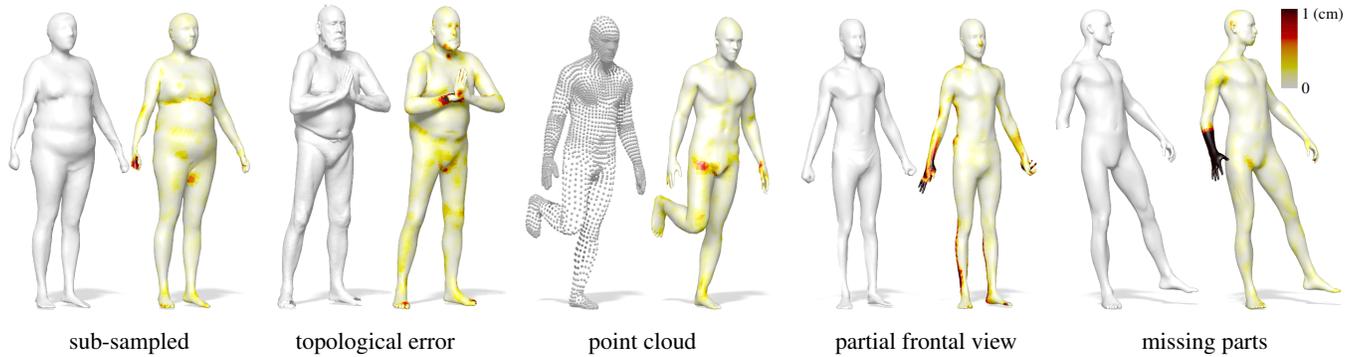

\subsection{Model fitting}
\label{sub:optimization}

\vspace{2ex}\noindent\textbf{Initialization.}
As a side-product of skeleton extraction, we have a functional map $T : L^2(\M)\to L^2(\N)$ at our disposal. This is converted into a pointwise map $\pi:\N\to\M$, and the resulting point-to-point matches are used, in turn, to estimate a {\em rigid} alignment among the two shapes.


\vspace{2ex}\noindent\textbf{Shape and pose regression.}
We now aim at bringing the template closer to the model by seeking for optimal shape and pose parameters. It should be noted that at this stage we do not seek yet for a perfect alignment, since this will be refined in  follow-up steps.
We minimize the following composite energy:
\begin{equation}
\mathcal{E} = w_S E_S + w_L E_L + w_{V} E_{V} + w_\beta E_{\beta} + w_\theta E_{\theta} 
\label{eq:eglobalpose}
\end{equation}
with respect to shape $\bm{\beta}$ and pose $\bm{\theta}$ (see Sec.~\ref{sub:SMPL}). Unless otherwise noted, for the rest of this Section we will tacitly assume that all quantities involved are functions of $\bm{\beta},\bm{\theta}$.

The $E_S$, $E_L$ and $E_{V}$ terms measure respectively the alignment error (in $\mathbb{R}^3$) of the skeleton joints, body landmarks, and surface vertices of the two shapes:
\begin{align}
E_S &=  \|\mathbf{S}_\M - \mathbf{S}_\N\|_F\,,\\
E_L &=  \|\mathbf{L}_\M - \mathbf{L}_\N\|_F\,,\\
E_{V}&=  \| \mathbf{X}_\M - \pi(\mathbf{X}_\N)\|_F 
\end{align}
where $\mathbf{S}_\M, \mathbf{L}_\M$ (resp. $\mathbf{S}_\N, \mathbf{L}_\N$) contain the 3D coordinates of skeleton joints and landmark positions for template and data shapes. Matrices $\mathbf{X}_\M,\mathbf{X}_\N$ contain the vertex coordinates for the two surfaces, and $\pi(\mathbf{X}_\N)$ denotes the image of points in $\N$ under the map $\pi$.
The terms
\begin{equation}
E_\beta =  \| \bm{\beta} \|^2\,,\quad\quad E_\theta = \mathbf{1}^\top\frac{\bm{\alpha}}{(\pi \mathbf{c}_\theta)^{12}}
\label{eq:ebeta}
\end{equation}
are regularizers for shape and pose (we only care about the rotation angles $\bm{\alpha}\in\mathbb{R}^{24}$ rather than the full transformations $\bm{\theta}$), to avoid the occurrence of very large values, and thus unrealistic body shapes and poses. Note that these regularization terms help to alleviate gross registration errors caused by a possibly noisy initial map. 
Here, division is meant element-wise and $\mathbf{c}_\theta \in \mathbb{R}^{24}$ is a constant vector specifying motion constraints for each of the 24 joints. We use the following values: 2 for joint 0 (max freedom of movement), $\frac{2}{18}$ for hands and feet, $\frac{5}{18}$ for body joints, $\frac{1}{36}$ for head and neck.

In our tests, we set the weights $w_S=10, w_L=1, w_{V}=0.1, w_\beta=0.5$. Minimization was performed using the dogleg method~\cite{dogleg} as implemented in the Chumpy automatic differentiation library~\cite{chumpy}.




\vspace{1ex}\noindent\textbf{Head and hands.}
At the end of the previous stage, the human template $\M$ is deformed in approximate alignment with the data $\N$. We now solve again problem \eqref{eq:comm} to obtain an improved functional map (note that the descriptors $f_i:\M\to\mathbb{R}$ are now computed on the {\em deformed} $\M$). This new map is used to obtain an improved skeleton for $\N$, and again to re-initialize the pose/shape regression step for the estimation of new model parameters for $\M$.

Differently from the previous stage, however, the energy \eqref{eq:eglobalpose} is modified with two additional terms that better constrain the alignment of head and hands (detected by growing geodesic balls around the corresponding landmarks).
The energy update is simply:
\begin{align}\label{eq:hh}
\mathcal{E}+
\| \mathbf{X}_\M^\mathrm{head}-\mathbf{X}_\N^\mathrm{head}\|_F+
\| \mathbf{X}_\M^\mathrm{hands}-\mathbf{X}_\N^\mathrm{hands}\|_F\,.
\end{align}
%



\vspace{2ex}\noindent\textbf{Non-rigid ICP.}
Since at this stage the deformed template is expected to align well with the data, we improve the registration further by alternating between the estimation of a point-to-point map $\pi_\mathrm{NN}$ via nearest-neighbor search in $\mathbb{R}^3$, and minimization of the bidirectional mean square error:
\begin{equation}
\| \mathbf{X}_\M - \pi_\mathrm{NN}(\mathbf{X}_\N)\|_F + \| \pi^{-1}_\mathrm{NN}(\mathbf{X}_\M) - \mathbf{X}_\N\|_F\,.
\label{eq:target2model}
\end{equation}
In the estimation of the map $\pi_\mathrm{NN}$, we filter out point-to-point pairings that have a large discrepancy (larger than $\frac{3\pi}{2}$) in the normal directions. Note, once again, that minimization of \eqref{eq:target2model} is done over shape and pose parameters $\bm{\beta},\bm{\theta}$.







\vspace{2ex}\noindent\textbf{Local refinement.}
Since the parametric model can only capture shape and pose within the span of its training set, an additional refinement step is required to reach a final, accurate registration. For example, the SMPL model (which we use in our experiments) does not capture head and hands articulations, while a model incorporating such details may not require refinement at this level. It is also important to note that, while artifacts are present when the hands are far from the default SMPL pose, they do not have a detrimental effect on the rest of the registration.

For the local refinement step, we employ an as-rigid-as-possible~\cite{Igarashi05} in conjunction with the nearest-neighbor energy \eqref{eq:target2model}.
%
%
However, differently from all previous steps, the vertex coordinates appearing in \eqref{eq:hh} are now optimized directly (i.e., they are not functions of $\bm{\beta},\bm{\theta}$).


%

\section{Registration results}
\label{sec:results}

\noindent\textbf{Data.}
In our experiments we use a wide selection of data collected from nine datasets exhibiting a variety of resolutions, sampling, surface artifacts and partiality. Specifically, we used: FAUST \cite{FAUST}, Princeton Segmentation Benchmark \cite{Chen:2009:ABF}, TOSCA \cite{TOSCA}, CAESAR \cite{robinette1999caesar}, KIDS \cite{rodola2014dense}, SHREC'11 \cite{SHREC2011}, SHREC'14 \cite{SHREC2014}, SPRING \cite{Yang2014} and K3D-hub \cite{Xu2017}.
All shapes were rescaled and downsampled to a similar density as the parametric model via edge collapse \cite{QSlim}, and small artifacts were fixed using MeshFix \cite{MeshFix}.

\begin{figure*}[t]
\begin{center}
\begin{overpic}
  [trim=0cm 0cm 0cm 0cm,clip,width=0.29\linewidth]{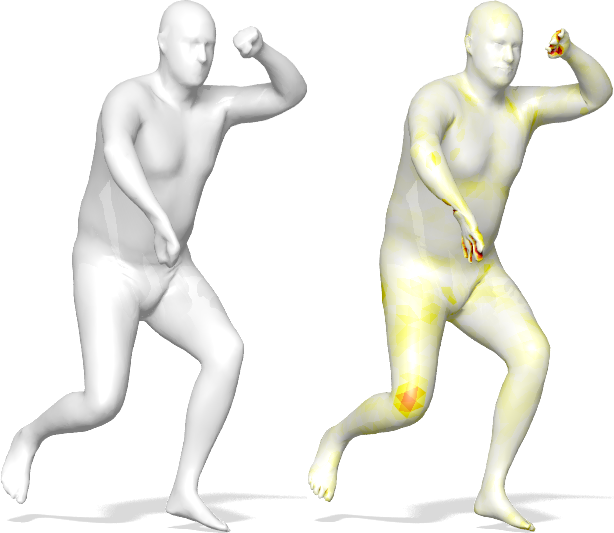}
\end{overpic}
\begin{overpic}
  [trim=0cm 0cm 0cm 0cm,clip,width=0.285\linewidth]{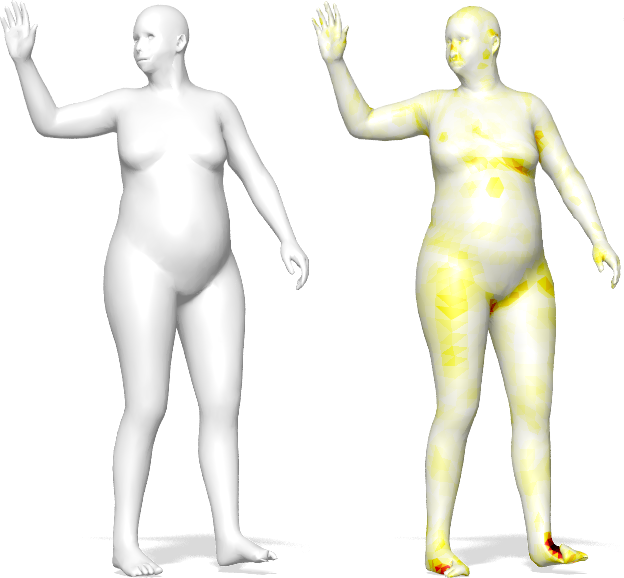}
\end{overpic}\hspace{0.23cm}
\begin{overpic}
  [trim=0cm 0cm 0cm 0cm,clip,width=0.17\linewidth]{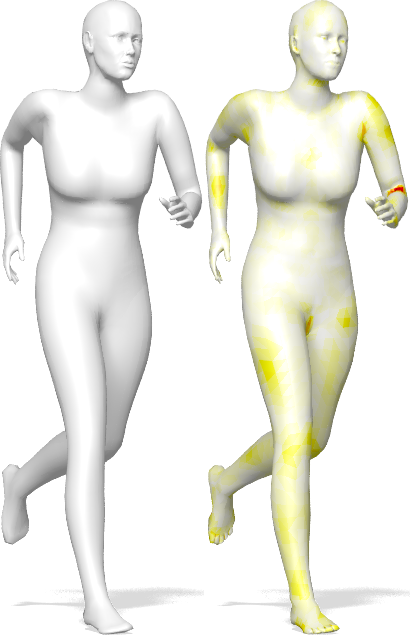}
\end{overpic}\hspace{0.23cm}
\begin{overpic}
  [trim=0cm 0cm 0cm 0cm,clip,width=0.22\linewidth]{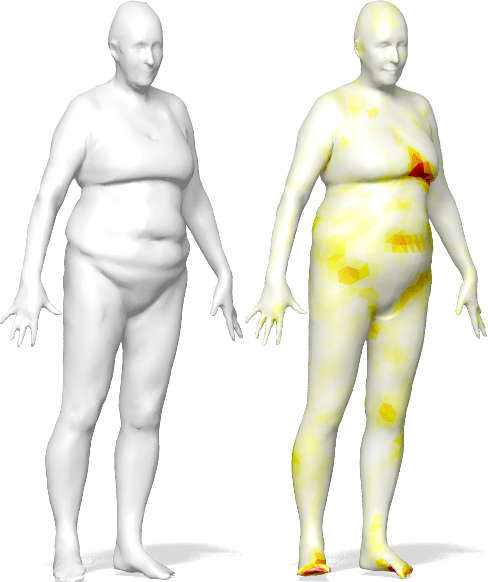}
\end{overpic}
\begin{overpic}
  [trim=0cm 0cm 0cm 0cm,clip,width=0.20\linewidth]{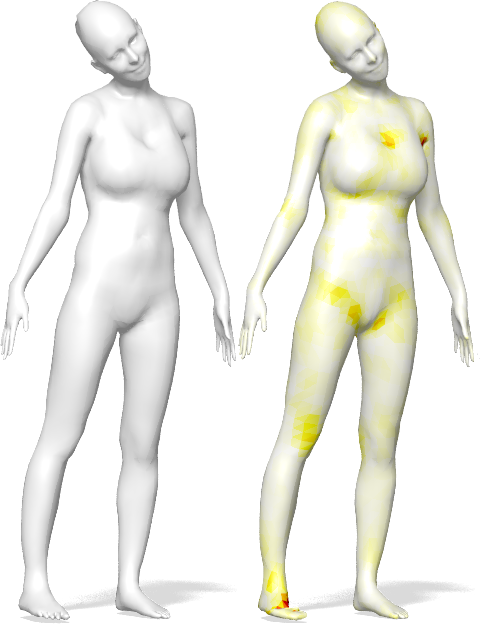}
\end{overpic}
\begin{overpic}
  [trim=0cm 0cm 0cm 0cm,clip,width=0.21\linewidth]{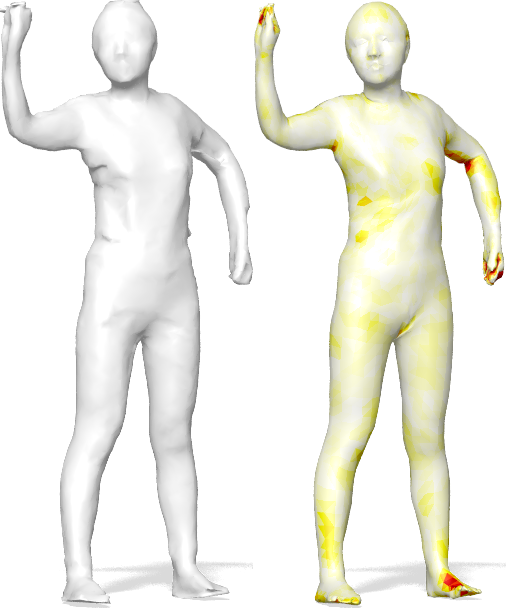}
\end{overpic}
\begin{overpic}
  [trim=0cm 0cm 0cm 0cm,clip,width=0.40\linewidth]{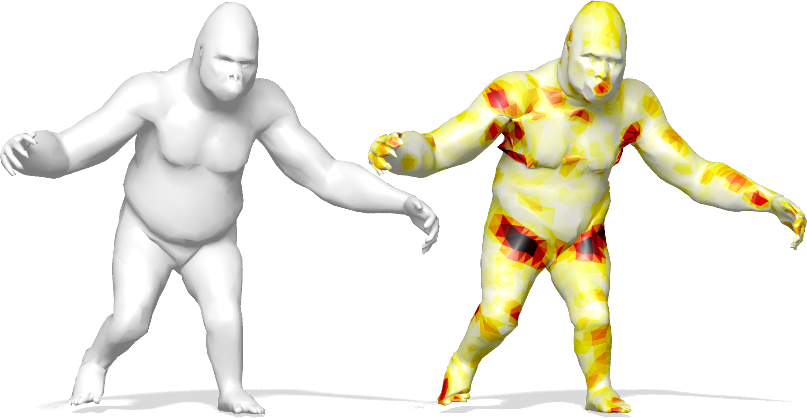}
\end{overpic}
\begin{overpic}
  [trim=0cm 0cm 0cm 0cm,clip,width=0.18\linewidth]{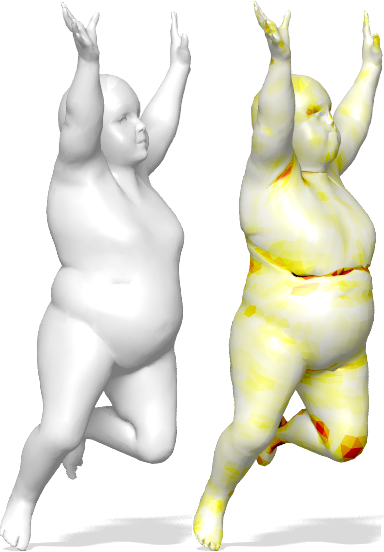}
\end{overpic}
\end{center}
\caption{\label{fig:regmore}Registration results for 8 shapes from different datasets. Note that the last two (gorilla and kid) are particularly challenging cases since they do not fall within the span of the underlying parametric model; the complete pipeline allows to obtain reasonable registrations for these cases as well.}
\end{figure*}

\vspace{1ex}\noindent\textbf{Robustness.}
We first evaluate the robustness of our pipeline under challenging perturbations. In Figure~\ref{fig:challenging} we show registration results under low resolution, topological error, point cloud representation, simulated rangemap, and missing parts respectively. Our pipeline achieves accurate results in all these cases; the registration error is close to zero almost everywhere, and otherwise smaller than 1cm. We refer to Figure~\ref{fig:regmore} and the supplementary material for additional results.

\begin{figure}[tb]
\begin{center}
\begin{overpic}
  [trim=0cm 0cm 0cm 0cm,clip,width=\linewidth]{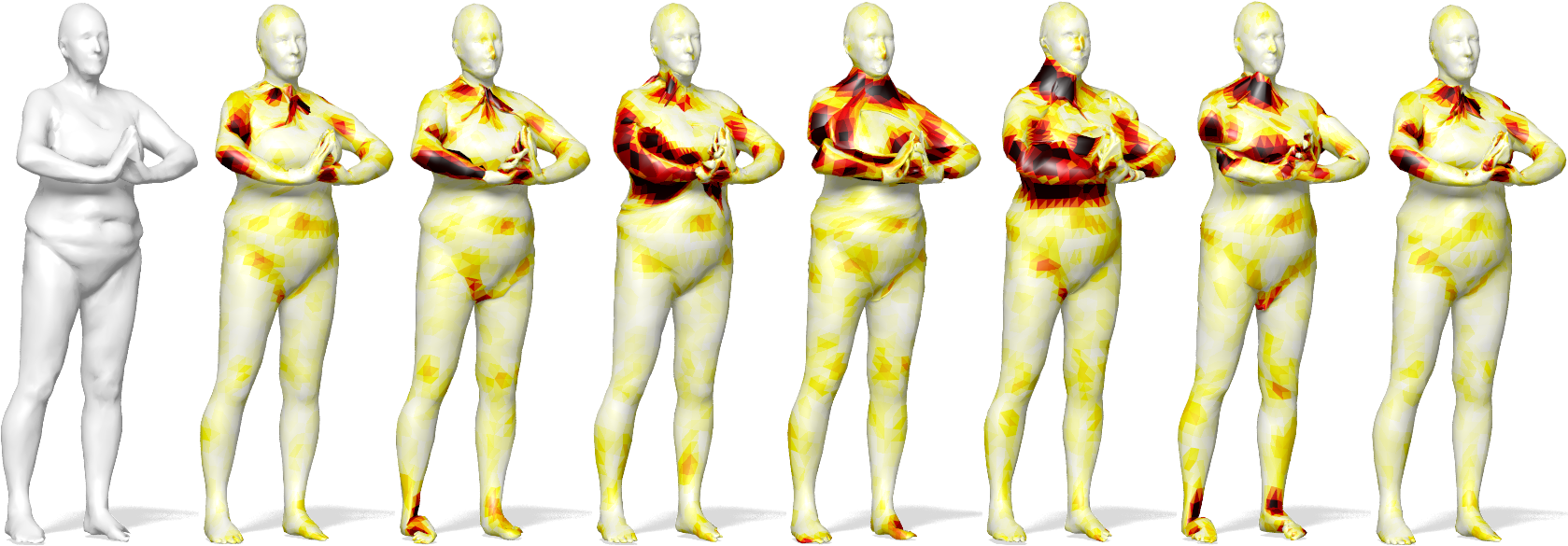}
\put(1, -3){\scriptsize Target}
\put(15, -3){\scriptsize \textbf{full}}
\put(12.5, -5){\scriptsize \textbf{pipeline}}
\put(26,-3){\scriptsize $w_{\beta}=0$}
\put(39,-3){\scriptsize $w_{S}=0$}
\put(49.5,-3){\scriptsize no normals}
\put(50.0,-5){\scriptsize constraints}
\put(62,-3){\scriptsize no head/}
\put(63.5,-5){\scriptsize hands}
\put(76.5,-3){\scriptsize $w_{\theta}=0$}
\put(88.5,-3){\scriptsize round 1}
\put(90,-5){\scriptsize only}
\end{overpic}
\end{center}

\vspace{0.25cm}

\caption{\label{fig:ablation}Ablation study. See main text for details.}
\end{figure}

\begin{figure}[tb]
\begin{center}
	\begin{overpic}
  [trim=0cm 0cm 0cm 0cm,clip,width=1\linewidth]{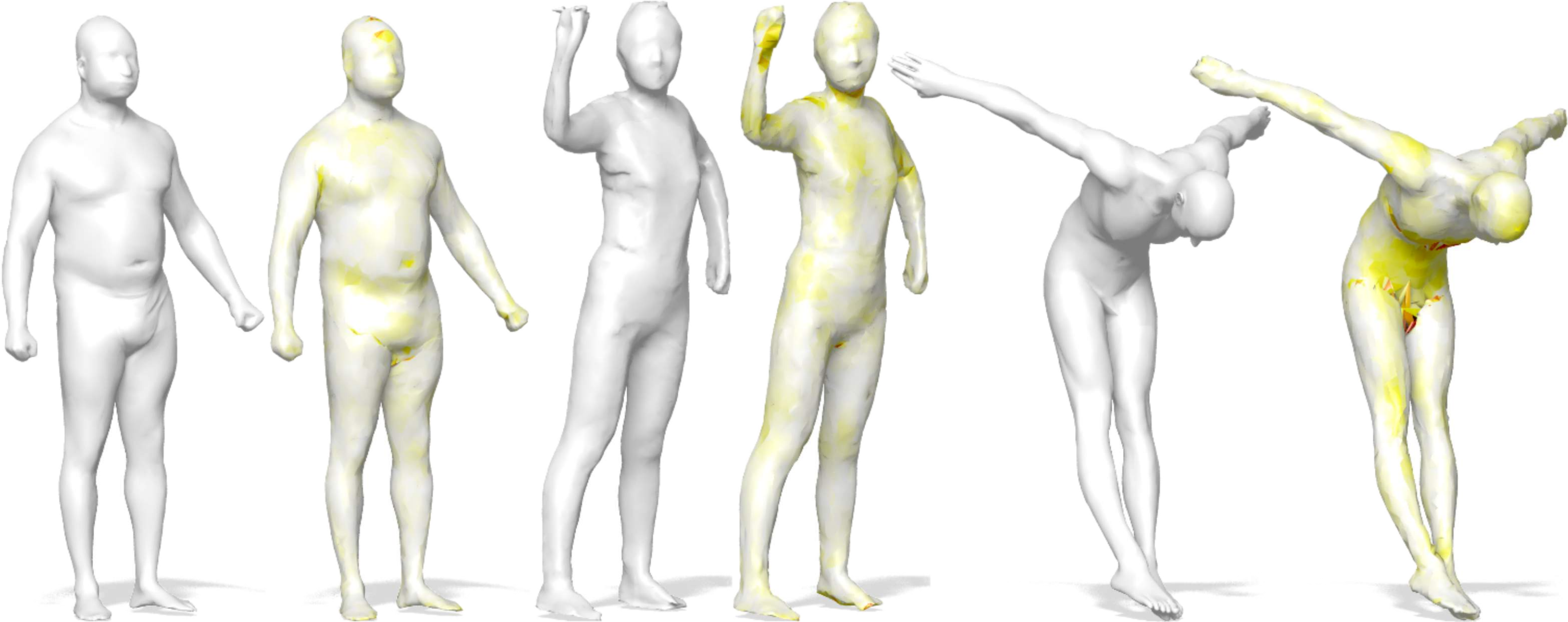}
\end{overpic}
	\caption{\label{fig:sscape}Registrations obtained by running our pipeline on top of the S-SCAPE parametric model. The other results in these pages employ the SMPL model.}
	\end{center}
\end{figure}

\vspace{2ex}\noindent\textbf{Ablation study.}
We conduct an ablation study in which the main terms of our composite energy \eqref{eq:eglobalpose} are disabled in turn, thus allowing to evaluate the effect of each within the registration process. Figure~\ref{fig:ablation} shows the results on a challenging case (with topological gluing).

Further, we show results as we change the underlying parametric model, namely by substituting SMPL with S-SCAPE~\cite{pishchulin15arxiv} without posture normalization. 
Within this model, pose is parametrized by 15 joints with the associated linear blend skinning weights. Since no joint regressor is provided, we define one by seeking for the minimizer:
\begin{equation}
\mathbf{R}^\ast = \arg\min_{\mathbf{R}\in\mathbb{R}^{15\times n_{\mathcal{M}}}} \| (\mathbf{W} \odot \mathbf{R} )\mathbf{X}_{\mathcal{M}} - \mathbf{S}_{\mathcal{M}} \|^{2}_F \,,
\label{eq:minjr}
\end{equation}
where $\mathbf{S}_{\mathcal{M}}$ contains the 3D joint coordinates of the S-SCAPE template.
The joint regressor is then defined by the element-wise product $\mathbf{W} \odot \mathbf{R}^\ast$, mapping surface vertices to skeleton joints.
%
%
%
The rest of the pipeline is applied as-is, yielding the results shown in Figure~\ref{fig:sscape}.

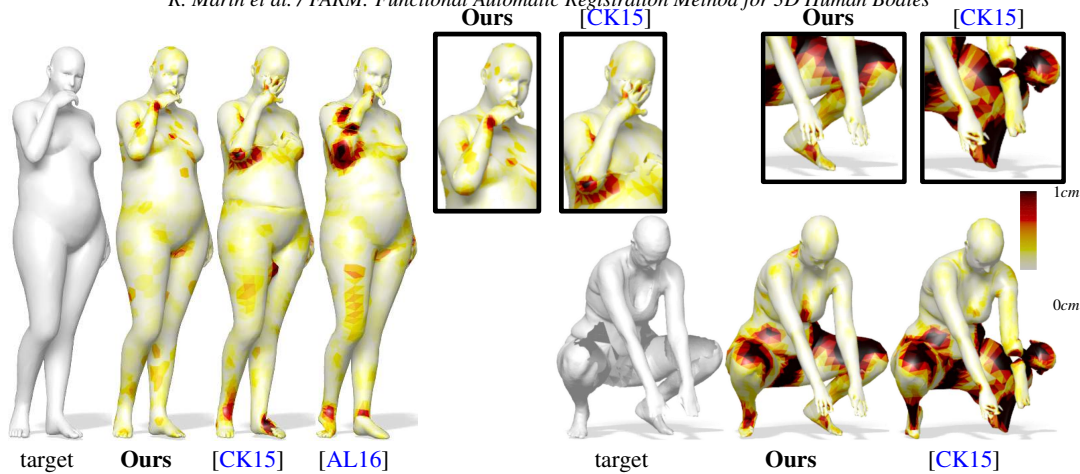
\begin{figure*}[t]
\begin{center}
\begin{overpic}
  [trim=0cm 0cm 0cm 0cm,clip,width=0.8\linewidth]{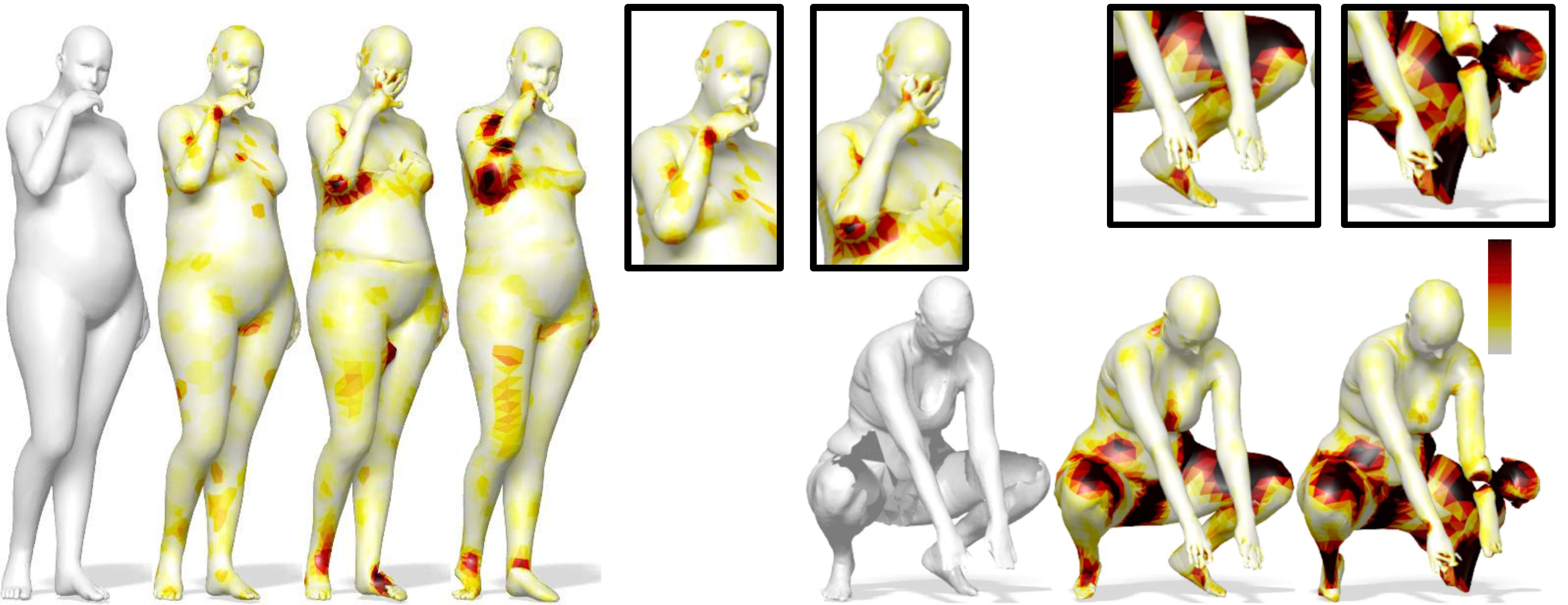}

 \put(1.1, -2.5){target}
 \put(10.5, -2.5){\thename}
 \put(18.5, -2.5){\cite{Koltun}}
 \put(28.5, -2.5){\cite{tutte}}

 \put(55, -2.5){target}
 \put(71, -2.5){\thename}
 \put(85.7, -2.5){\cite{Koltun}}

 \put(42.5, 39){\thename}
 \put(53, 39){\cite{Koltun}}
 \put(74.5, 39){\thename}
 \put(88, 39){\cite{Koltun}}

\put(98, 22.8){\scriptsize $1cm$}
\put(98, 12.2){\scriptsize   $0cm$}
\end{overpic}
\end{center}
\vspace{0.3cm}
\caption{\label{fig:comparison}Registration performance when our correspondence step is replaced with the matching pipelines of \cite{Koltun} and \cite{tutte}. Note that \cite{tutte} can not be applied on shapes with different genus. The black regions on the legs of the right example are due to the part being completely missing, as it can be seen from the target (i.e., it is not due to registration error).}
\end{figure*}

Finally, our pipeline involves a map inference step that can be substituted with other matching approaches. We therefore adopt the matching pipelines \cite{Koltun} and \cite{tutte} as a plug-in replacement for our correspondence estimation step, while keeping the other steps of the pipeline unchanged.
In particular, \cite{Koltun} is among the state of the art for shape matching as evaluated on the FAUST challenge \cite{FAUST}; \cite{tutte} is the only method giving guaranteed continuous bijections, but requires a sparse input correspondence (we use the 5 landmarks) and does not minimize metric distortion. The results are shown in Figure~\ref{fig:comparison}, highlighting the effectiveness of our entire pipeline. We refer to the supplementary material for additional examples.

%

%

\section{Applications}
\label{sec:applications}

We finally showcase our registration method in three different applications.

\subsection{Shape correspondence}
Shape registration provides point-to-point correspondences among the involved shapes as a side product. We therefore evaluate our pipeline for this task on the FAUST benchmark \cite{FAUST}, consisting of real scans of human subjects acquired using a full-body 3D stereo capture system. These scans exhibit geometric noise, topological errors, and missing parts. The ground-truth correspondences for the challenge are {\em not} provided, rather an accuracy evaluation is obtained by submitting correspondence results online. 

\begin{figure}[t]
\begin{center}
	\includegraphics[width=\columnwidth]{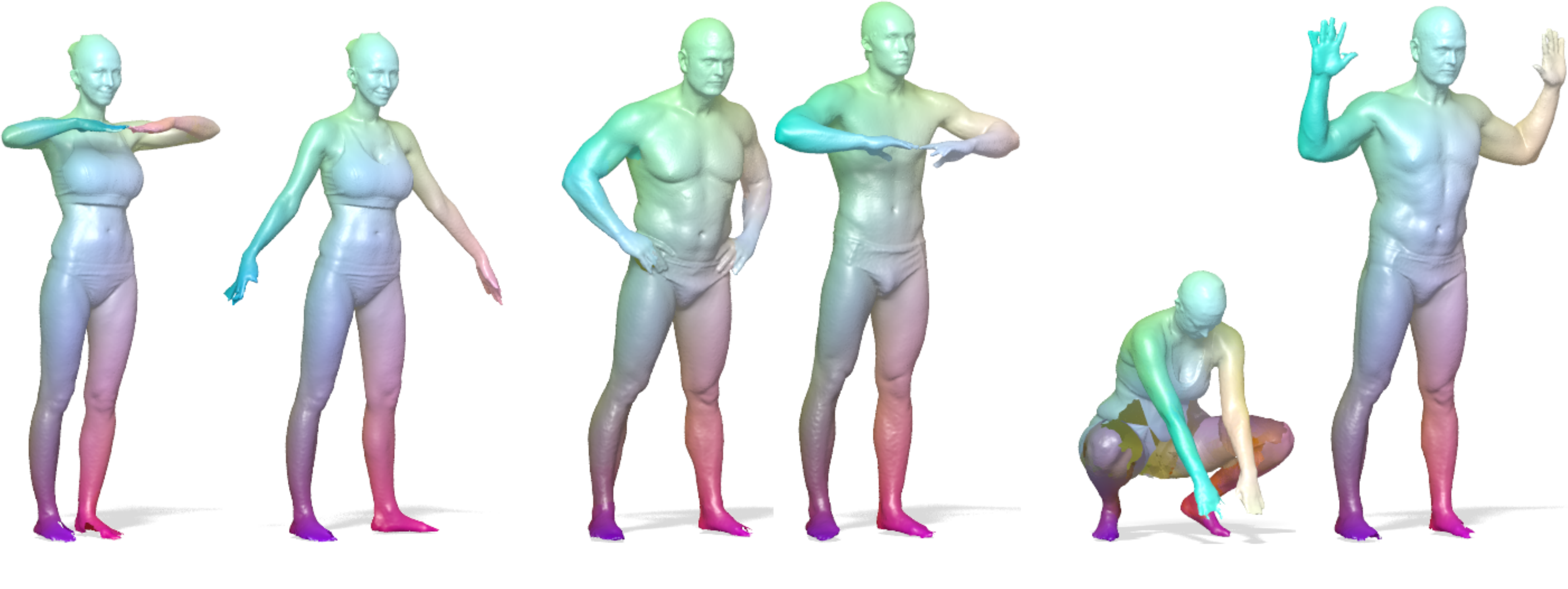}
	\includegraphics[width=0.29\columnwidth]{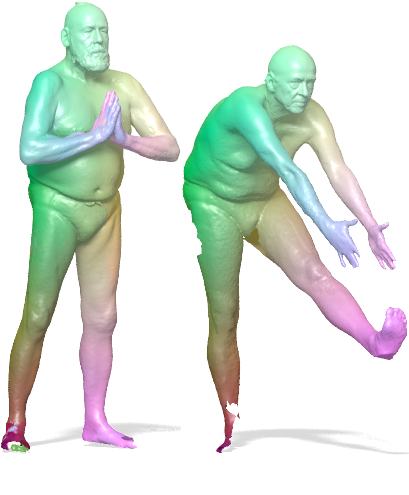}
	\hspace{0.1cm}
	\includegraphics[width=0.33\columnwidth]{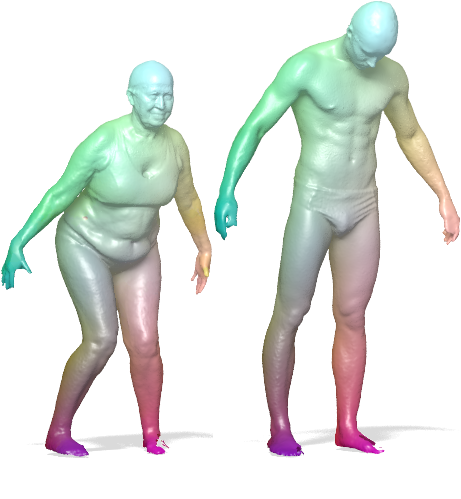}
	\hspace{0.1cm}
	\includegraphics[width=0.32\columnwidth]{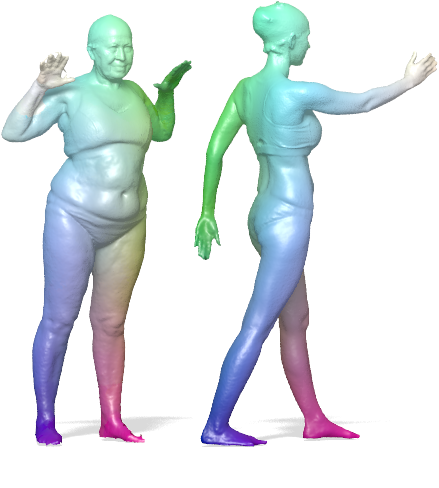}
	\vspace{-0.7cm}
	\caption{\label{fig:FAUST}Dense correspondence results on six pairs from the FAUST challenge (real scans with $\sim$350K triangular faces). These cases include pose change, different subjects, missing geometry and mesh gluing. Corresponding points are visualized with same color.}
	\end{center}
\end{figure}

Given a challenge pair, we apply our registration pipeline to each of the two shapes individually. Once the parametric model is registered to the two shapes, we are able to establish point-to-point correspondences via this common domain and then pull them back to the original meshes. Correspondences obtained this way are used to initialize a matching step according to \eqref{eq:refine}.
Examples of matching results are shown in Figure~\ref{fig:FAUST}, and a quantitative comparison with the official ranking is reported in Table~\ref{tab:FAUST}. We refer to the supplementary material for a complete pair-by-pair evaluation.

%
%
\begin{table}[t]
\centering
\resizebox{\linewidth}{!}{
\begin{tabular}{|l|c|c|c|c|}	
\hline
 & \textbf{\shortstack{Inter AE}} & \textbf{\shortstack{Inter WE}}& \textbf{\shortstack{Intra AE}} & \textbf{\shortstack{Intra WE}} \\\hline
Zuffi et al. \cite{SPuppet} & 3.13  & 6.68 & 1.57 &  5.58 \\
Chen et al. \cite{Koltun} & 8.30 & 26.80 &  4.86 &  26.57 \\
Litany et al. \cite{FMNET} & 4.83  & 9.56 & 2.44 &  26.16 \\
Fan et al. \cite{CHARM} & n.a. & n.a. & 15.16 & 57.14 \\								\textbf{FARM} (Ours) & 4.12 & 9.98 &  2.81 & 19.42 \\\hline
\end{tabular} 
}
\caption{\label{tab:FAUST}Comparison on the FAUST challenge. {\em AE} and {\em WE} denote average and worst error respectively.}
\end{table}

\subsection{Shape completion}
As another application, we consider completion of partial deformable 3D shapes. To illustrate the flexibility of the registration pipeline, we look at both synthetic (artificial cropping of clean meshes) and real-world (incomplete Kinect and D-FAUST~\cite{dfaust} scans) data; we stress that the pipeline is applied as-is in all cases, with no further adjustments or tuning to account for the challenging setting. 

\begin{figure}
\begin{center}
	\includegraphics[width=\columnwidth]{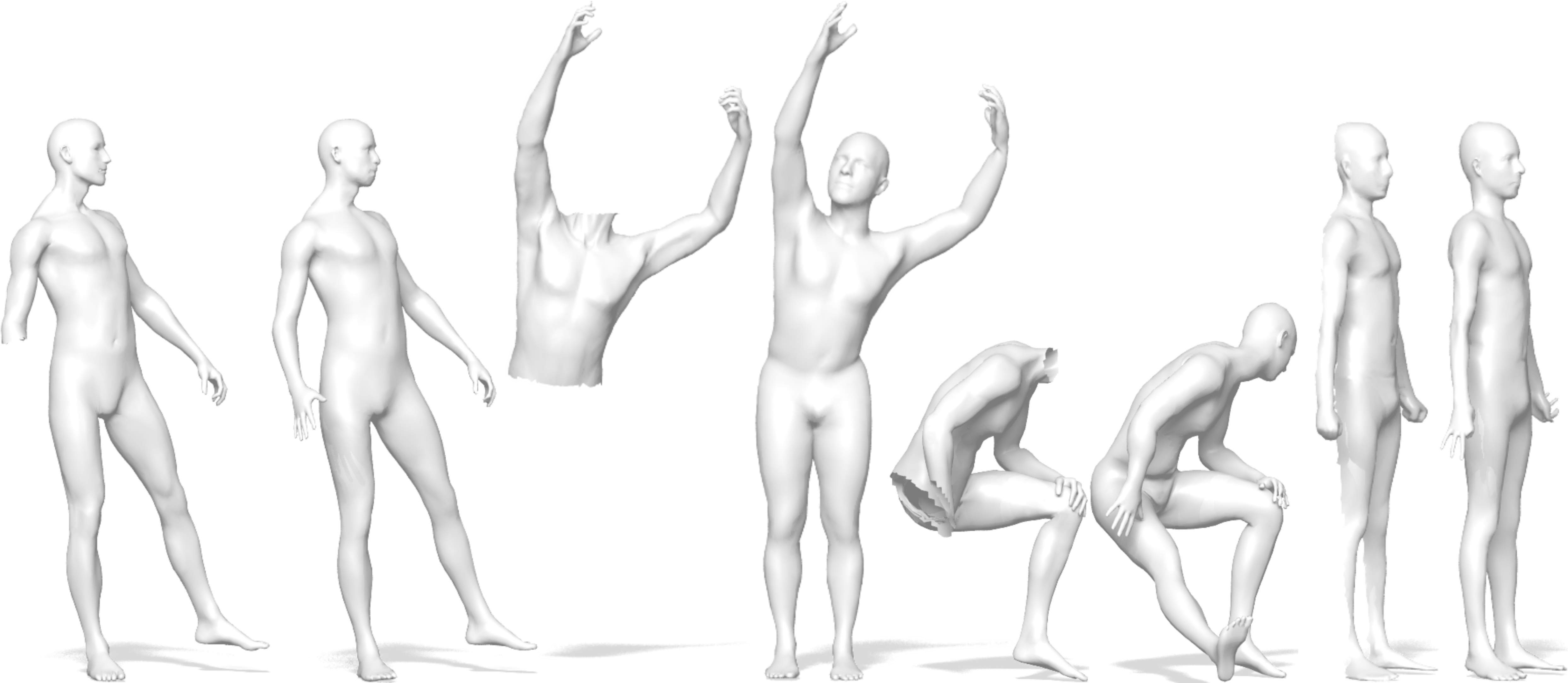}
	\caption{\label{fig:completion}Deformable shape completion results. For each pair, we show the incomplete input (left) and the completed mesh (right).}
	\end{center}
\end{figure}

\begin{figure*}[h]
\begin{center}
	\begin{overpic}
  [trim=0cm 0cm 0cm 0cm,clip,width=1\linewidth]{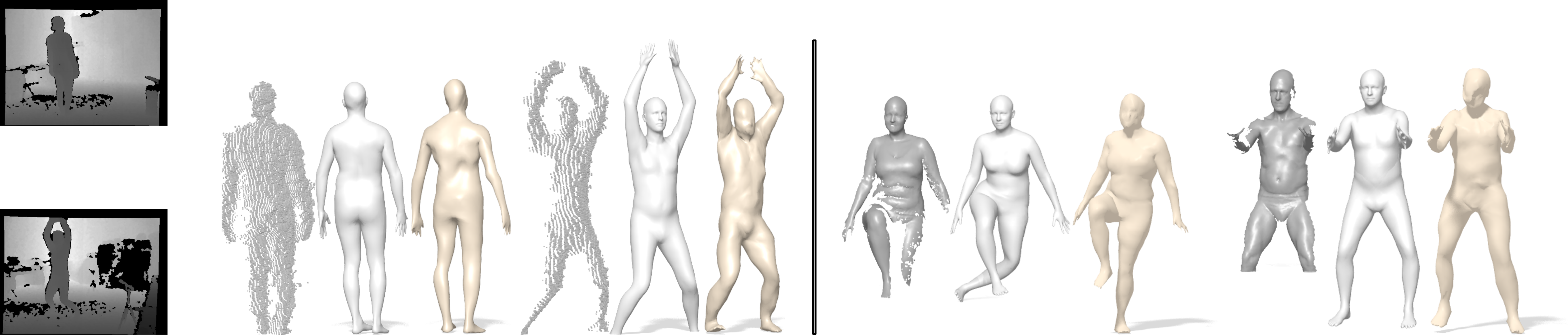}

	\put(4.5, -2){(b)}
	\put(4.5, 11.3){(a)}
	
\put(18.2, 20){Kinect scan (a)}
 \put(37.2,20){Kinect scan (b)}
\put(59.3, 20){DFAUST scan}
 \put(83.5, 20){DFAUST scan}

\put(21.5, -3){\thename}
 \put(26.4, -3){\small \cite{completion}}
\put(40.0, -3){\thename}
 \put(44.5, -3){\small \cite{completion}}
\put(61.4, -3){\thename}
 \put(67.9, -3){\small \cite{completion}}
\put(85.6, -3){\thename}
 \put(90.6, -3){\small \cite{completion}}

\end{overpic}
\vspace{0.1cm}
	\caption{\label{fig:rangemap}Deformable shape completion with real scans. We compare with the deep learning method of Litany et al.~\cite{completion}, currently the state of the art method for this task.}
	\end{center}
\end{figure*}

Results on synthetic data are reported in Figure~\ref{fig:completion}, while in Figure~\ref{fig:rangemap} we compare with the state-of-the-art deformable shape completion method of Litany et al.~\cite{completion}. Note that the latter method adopts a fully supervised deep learning model (graph convolutional autoencoders), and is limited in terms of mesh resolution. In all these experiments, we let our parametric model assume default parameter values at the joints for the shape parts that do not have a corresponding region in the input data (these are detected automatically during the matching step).
%
%


\subsection{Shape modeling and animation}
\begin{figure}
\begin{center}
\begin{overpic}
[trim=0cm 0cm 0cm 0cm,clip,width=\columnwidth]{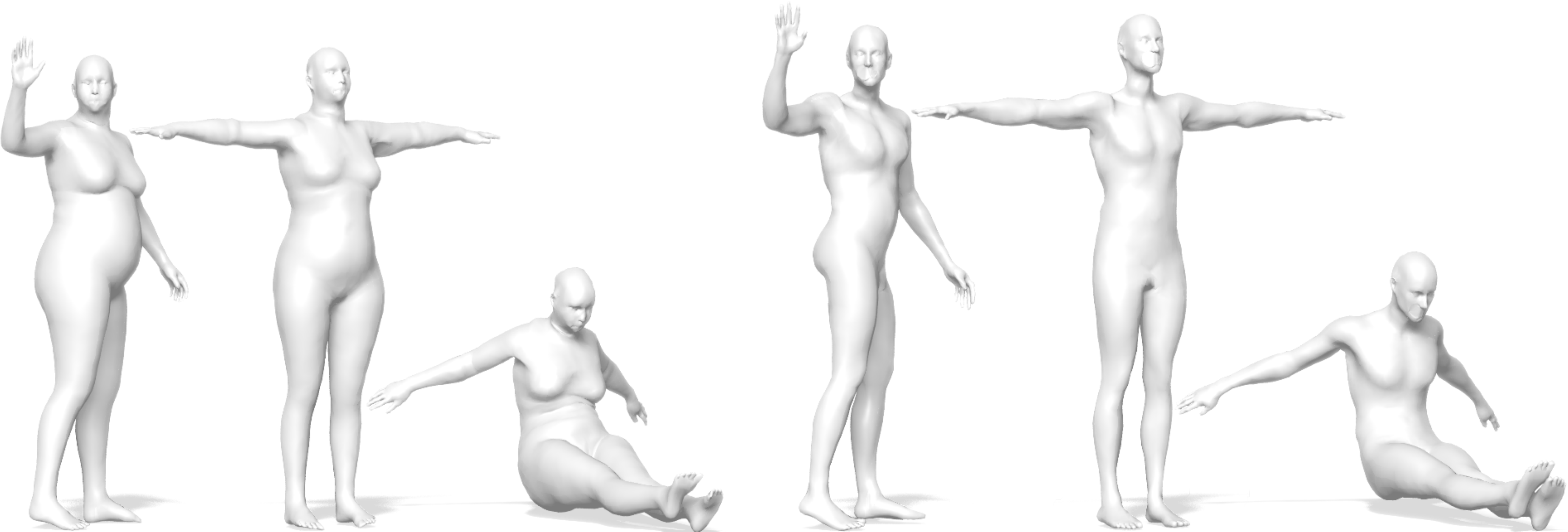}
\put(3,-3.7){\footnotesize $\mathcal{N}_1$}
\put(16,-3.7){\footnotesize T-pose}
\put(27.5,-3.7){\footnotesize pose from $\mathcal{N}_2$}
\put(46,5){\line(0,1){24}}
\put(47,-3.7){\footnotesize pose from $\mathcal{N}_1$}
\put(69,-3.7){\footnotesize T-pose}
\put(90,-3.7){\footnotesize $\mathcal{N}_2$}
\end{overpic}

\vspace{0.35cm}

	\includegraphics[width=\columnwidth]{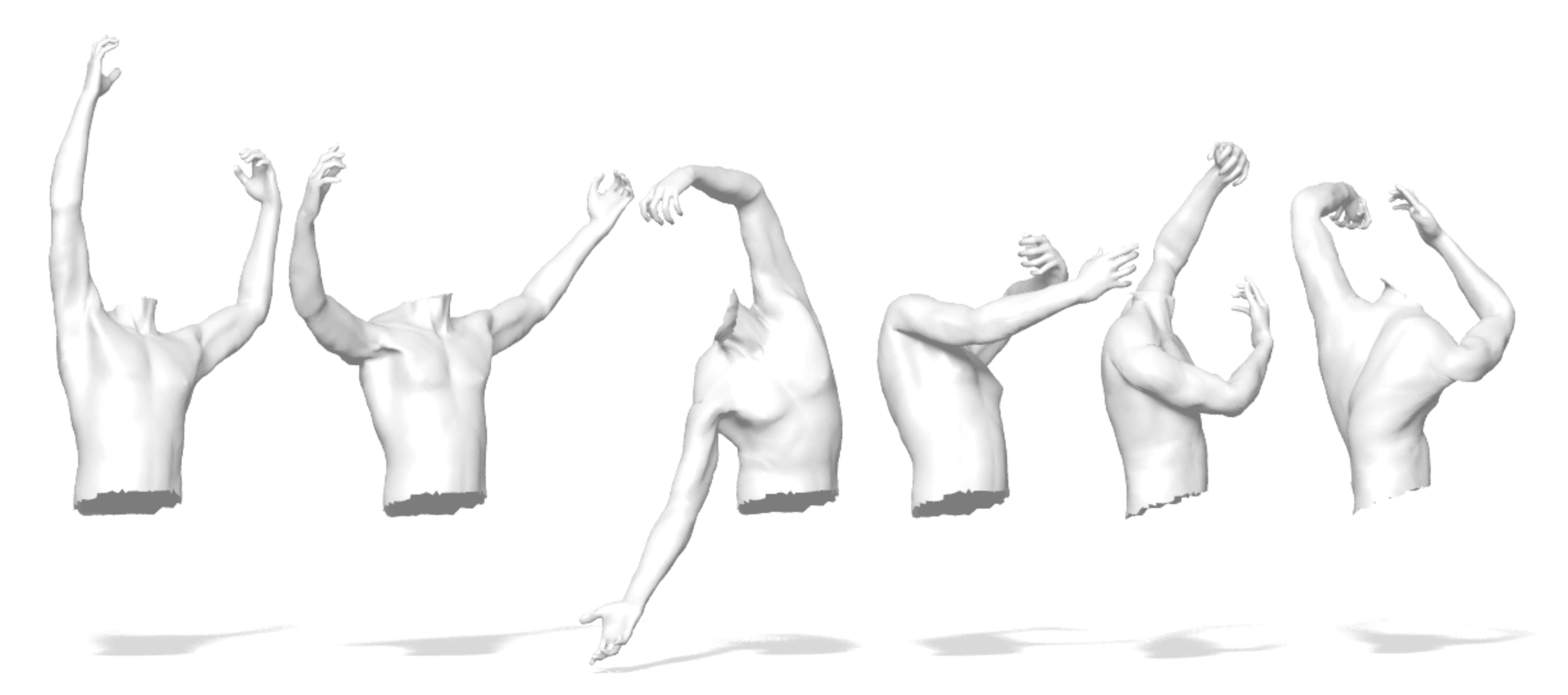}
	
	\vspace{-0.2cm}
	
	\caption{\label{fig:animation}{\em Top}: Transferring pose between two full shapes. {\em Bottom}: Transferring skinning information across partial shapes.}
	\end{center}
\end{figure}

Finally, we showcase the application of our registration method in a character animation pipeline. Once the parametric model is registered to the data, the skinning information is transfered to the latter and one can ``undo'' the data shape to a T-pose. From here, motion parameters can be applied to animate the character or transfer animations across multiple shapes. See Figure~\ref{fig:animation} for examples on full and partial data; we refer to the supplementary material for additional details and more examples.
%
%
%

\section{Conclusion}
\label{sec:conclusion}
We presented a novel approach for the fully automatic registration of non-rigid human shapes. 
The main \textbf{limitations} of our method are to be found in its direct dependence on the underlying parametric model, which ultimately determines the quality of the final alignment as demonstrated in dedicated tests.
%
%
A particularly interesting direction for future work is the introduction of localized manifold harmonics \cite{LMH} in the map inference steps, which would enable the application of our method in the presence of cluttered scenes \cite{cosmo2016matching} without any supervision.
 


\section*{Acknowledgments}
We thankfully acknowledge Silvia Zuffi, Federica Bogo and Maks Ovsjanikov for their valuable comments and the technical support. ER is supported by the ERC grant no. 802554 (SPECGEO).


\bibliographystyle{eg-alpha-doi}

\bibliography{egbib_Simo}
\end{document}